%% file: main.tex
\documentclass[3p, twocolumn]{elsarticle}
\usepackage[utf8]{inputenc}
\usepackage{comment}
\usepackage[colorlinks,linkcolor=blue,citecolor=blue,urlcolor=blue]{hyperref}
\usepackage{amsmath}
\usepackage{amsfonts}
\usepackage{amssymb}
\usepackage{hyperref}
\usepackage{cleveref}
\usepackage{cuted}
\usepackage{lipsum}
\usepackage{pgf}
\usepackage{todonotes}

\makeatletter
\def\ps@pprintTitle{%
 \let\@oddhead\@empty
 \let\@evenhead\@empty
 \def\@oddfoot{}%
 \let\@evenfoot\@oddfoot}
\makeatother

\makeatletter
\def\@author#1{\g@addto@macro\elsauthors{\normalsize%
    \def\baselinestretch{1}%
    \upshape\authorsep#1\unskip\textsuperscript{%
      \ifx\@fnmark\@empty\else\unskip\sep\@fnmark\let\sep=,\fi
      \ifx\@corref\@empty\else\unskip\sep\@corref\let\sep=,\fi
      }%
    \def\authorsep{\unskip,\space}%
    \global\let\@fnmark\@empty
    \global\let\@corref\@empty  
    \global\let\sep\@empty}%
    \@eadauthor={#1}
}
\makeatother

\title{DBCal: Density Based Calibration of classifier predictions for uncertainty quantification}
\author[pnnl]{Alex Hagen\corref{cor1}}
\cortext[cor1]{alexander.hagen@pnnl.gov}
\author[pnnl]{Karl Pazdernik}
\author[pnnl]{Nicole LaHaye}
\author[pnnl]{Marjolein Oostrom}
\address[pnnl]{Pacific Northwest National Laboratory, Richland, WA, USA}
\date{February 2022}

\begin{document}

\begin{abstract}
    Measurement of uncertainty of predictions from machine learning methods is important across scientific domains and applications.  We present, to our knowledge, the first such technique that quantifies the uncertainty of predictions from a classifier and accounts for both the classifier's \emph{belief} and \emph{performance}.  We prove that our method provides an accurate estimate of the probability that the outputs of two neural networks are correct by showing an expected calibration error of less than 0.2\% on a binary classifier, and less than 3\% on a semantic segmentation network with extreme class imbalance.  We empirically show that the uncertainty returned by our method is an accurate measurement of the probability that the classifier's prediction is correct and, therefore has broad utility in uncertainty propagation.
\end{abstract}

\maketitle

\section{Introduction}

Many scientific tasks can be augmented with the predictive power of neural networks \cite{Hagen2021, Kautz2019, Pazdernik2020, Parsons2020}, however a consensus does not exist on best practices to quantifying the uncertainty in decisions and results stemming from these neural networks \cite{gawlikowski_survey_2022, Castelvecchi2016}. Scientific rigor and subsequent justification of decisions requires proper uncertainty justification; this applies across fields such as nuclear science (as is presented as an example in this work) \cite{Pazdernik2020}, medical diagnostics \cite{Tian2020}, or forensics \cite{Burr2021}.

As we will show in subsequent sections, no modern uncertainty quantification method can provide all of the desiderata of physical scientists' and other analysis consumers. Variational methods require increased computational cost compared to traditional inference, and often are not statistically calibrated: we show that the reported uncertainty of common variational methods is indicative of the fragility of a model to perturbation, not a quantification of the uncertainty of the model with respect to the ground truth. Architecturally based methods require stipulations on the model architectures possible, and often have less expressive power because they are parametric.  Finally, no method can incorporate both the model's internal uncertainty about its predictions, and the uncertainty apparent from the model's performance on a validation set, regardless of its internal confidence.

We present two novelties in this work.  The first is an interpretation of the requirements of an uncertainty estimate for neural network predictions to incorporate performance, the second is a method to realize all of these requirements. Our method is simple, and quantifies the uncertainty of any classifier with continuous real number output (such as the logits layer of neural networks) by statistically calibrating that quantification compared to the ground truth. While the classifier may be considered parametric, our uncertainty quantification is non-parametric and thus exhibits high expressive power.  The method benefits from variational methods to increase the number of samples upon which it can be calibrated, but does not require them.  The method connects the model's internal confidence mechanism with the model's empirical performance, which has a side benefit of correcting neural network's often cited ``overconfidence" problem \cite{Hsu2020}.  While the method is more computationally costly than training a neural network without uncertainty quantification, it is a constant cost addition applied after training, unlike the linear cost addition (with respect to the number of training iterations) added by other variational methods.

We present in the following sections a survey of existing quantification methods and point out benefits of each.  We describe the gap between current variational inference methods and the desired useful and calibrated uncertainty. Then, we formally present our method. Subsequently, we provide an example of our method on a toy problem: that of a binary classifier trained on a balanced data set.  We then apply our method to semantic segmentation of micrographs of Lithium Aluminate, an application important in nuclear science; this example shows the method's capability to perform on multiclass classification as well as an unbalanced data set. We conclude with a summary of this methodology and discussion on future work.

\subsection{Relevant literature}

The lack of uncertainty quantification is an oft cited \cite{Castelvecchi2016} detriment to deep learning, due to deep learning's ``black box" nature.  That weakness has lead to a sizeable literature in uncertainty quantification (UQ) for neural networks (NN). In our review, we found that methods for UQ are split into two main subclasses: uncertainty aware neural network architectures, such as Bayesian neural networks \cite{Hernandez-Lobato2015, izmailov_what_2021, hoffmann_deep_2021}; and variational uncertainty quantification, as in Bayesian SegNets \cite{Kendall2015, gawlikowski_survey_2022}. We focus on the second subclass because of its flexibility: one can theoretically apply it to any architecture of neural network.  We also note that, while modifications of architectures to quantify uncertainty by design have achieved notable results, the statistics underlying continuous prediction scores is a still-developing field.  Both the continuous Bernoulli distribution \cite{Loaiza-Ganem} and continuous categorical distributions \cite{Gordon-Rodriguez2020} were unpublished at the time of the design of Bayesian neural networks.

While high level discussions of UQ have helped inspire this work \cite{Gustafsson2020}, our method is conceptually most similar to the ``Learn by Calibrating" (LbC) method \cite{Thiagarajan2020}.  Our application, and initial investigations, were most similar to Bayesian SegNet (BSN) \cite{Kendall2015}, although our method eventually diverged from their method, due to the ambiguous calibration of BSN.

\subsection{Variational inference is not enough}

\begin{figure}
    \centering
    \includegraphics{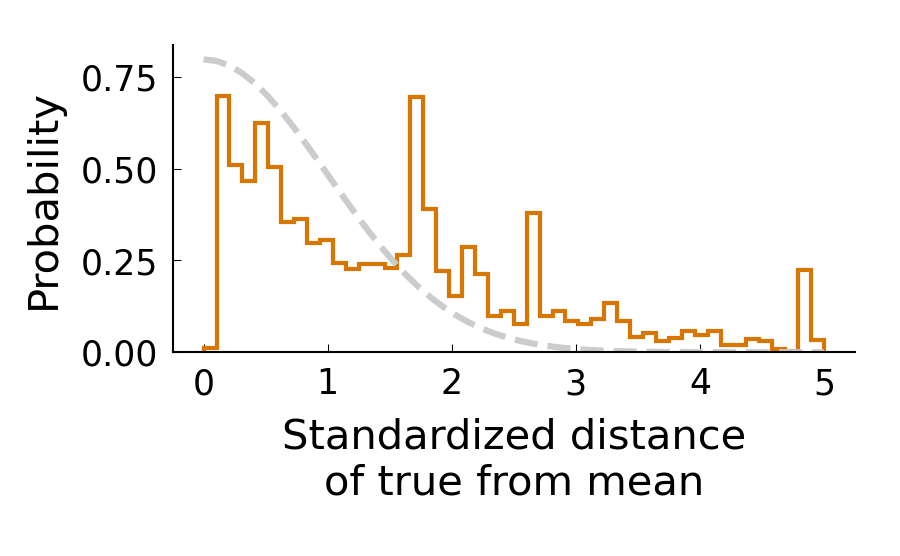}
    \caption{Prediction and variance of prediction using Bayesian SegNet method \cite{Kendall2015}. We find that the variance of predictions are not well calibrated: The distribution of scores is clearly not normally distributed. Only 46.7\% of predictions fall within 1 standard deviation of their true values, and there is a peak at almost 2 standard deviations from the true value due to the model's confusion between classes.}
    \label{fig:bsn_bad}
\end{figure}

The BSN and similar variational inference methods quantify uncertainty by using Monte Carlo Dropout layers in the neural network during both training and inference. This layer adds variance into the output of the neural network.  Kendall et al. published a popular strategy for variational inference, which we discuss and compare our method to in this section.  They suggest performing several "trials" with the same inputs and recording the distribution of the output predictions.  The variance of those outputs is computed and then used as the uncertainty of the prediction.  We found that this variance does not provide a well calibrated estimate of uncertainty. While it is not specified, this variance is then often used with the assumption of normality. We show that the raw neural network outputs, standardized by the variance, do not form a normal distribution around their true value, and that confusion between classes in the model inflates that variance because of ``peaks" in the standardized neural network output distribution. This is shown in \cref{fig:bsn_bad} for the imbalanced classification (semantic segmentation) problem described in \cref{sec:lialo2}.

More fundamentally, in the multiple class case, it is ambiguous what the variance in a score means.  If the value in dimension 0 has mean 0.9 and standard deviation 0.46, it is unclear whether a prediction one standard deviation away from the mean corresponds to all of that score moving to a single other class (which would result in a change of the argmax class prediction), or spread throughout all other classes (which would not result in a change of the argmax class prediction). Nor is this variance rigorously connected to the ground truth.

Instead, we desire a technique that, given a neural network and an input datum, provides a predicted probability of the prediction being correct. We point out that this desired method must necessarily be connected to the model's \emph{performance}, as well as its internal \emph{belief}; this is where we diverge from over methods.  We define the model's belief to be that which others measure: the magnitude and variance of a network's raw score given each input. We define the model's performance to be the accuracy with which its predictions match the true class.  For a two-class case, it is clear that a model which predicts $\left[0.9, 0.1\right]$ is more confident in its prediction of class 0 than one which predicts $\left[0.51, 0.49\right]$. This does not evaluate the probability of that prediction being correct.

An (admittedly pathological) thought experiment of switching the labels after training illustrates that measuring the model's belief can be completely divorced from the the model's performance. Drawing from other computational sciences, we posit that current uncertainty quantification techniques are more similar to ``verification" or ``perturbation analyses" \cite{Thacker2004}, in that they measure values proportional to the likelihood that a classifier's true prediction is the same as the current prediction, given the same input many times, \emph{regardless whether that prediction is the same as the ground truth or not}. For physical sciences, a measurement akin to ``validation" is more useful, therefore we believe the uncertainty of a model should be related to the likelihood that a classifier's prediction is the true value. Therefore, we connect the uncertainty prediction to both the model's belief and performance. We now formalize this method in \cref{sec:calibration}.

\section{Calibration}\label{sec:calibration}

Our calibration method is conceptually simple.  We view the output of the neural network (whether logits or softmax) for some classification as a point in space, and simply compare the density of correct and all classifications local to that point in space. The formalization of that procedure follows.

We have a $d$ class classifier $y=f\left(x\right)$ where $y\in \mathbb{R}^d$.  We also know the true classification of a validation set $C_{val}$ given inputs $x_{val}$.  We desire to know $y_{val}=f\left(x_{val}\right)$, the predicted class labels of the inputs, and $p_{val}$, the probability that the predicted class is correct.

Knowing $y_{val}$, we can determine the discrete predicted class $c_{val}$ using the argmax function.  Now, we separate $y_{val}$ into partitions where the predictions were correct, $y_{correct}$, by comparing $c_{val}$ and $C_{val}$.

While neural networks often display overconfidence in their predictions \cite{Hsu2020}, the likelihood of a prediction being correct has been empirically shown to be related to the maximum value of the score.  We take this as axiomatic to build the rest of our arguments.  We posit that the neural network will predict similar scores for similar data points and that those similar data points have similar likelihood of being of the same class.  Therefore, we calculate the local density of correct scores and compare that to the density of all scores local to that point in order to generate an estimate of the current prediction's probability of being correct. Formally, we posit that the scores $y_{val}$ are a random variate with distribution $\mathcal{Y}$ such that $y\sim\mathcal{Y}$. We measure the probability density of that distribution with some density estimate, $\rho$.

With the density estimates described above, the probability density
\begin{equation}
\rho_{c}\left(y\right)\equiv\rho_{\mathcal{Y}}\left(y|c_{val}=C_{val}\right)
\end{equation}
and
\begin{equation}
\rho_{a}\left(y\right)\equiv\rho_{\mathcal{Y}}\left(y\right)
\end{equation}for correct and the overall density, respectively. We can calculate the overall probability of being correct, given score $y_{new}$, as
\begin{equation}\label{eq:prob_correct}
    p\left(y_{new}\right) = \frac{\rho_{c}\left(y_{new}\right)}{\rho_{a}\left(y_{new}\right)}
\end{equation}.

We note several interesting points.
\begin{itemize}
    \item Our method does not require variational inference, although it does require enough data to enable an accurate density estimate in a space with $d$ dimensions. Our method is compatible with variational inference, which we discuss below.
    \item Our method can be used no matter the training status of the neural network. In fact, we show below that it provides calibrated estimates even when the model is poorly trained.
    \item Density estimation can always be performed in $d$ dimensions. When the output of the neural network is modified by a softmax layer, the density estimation can be performed in $d-1$ dimensions, as any vector undergoing the softmax operation subsequently exists on a simplex in $d$ dimensions, which is reducible to $d-1$ dimensions.  We have noticed no difference in calibration between the two methods.
    \item This method requires careful use of validation data.  In general, we desire that any evaluation of the calibration be performed on data that was used neither to train the network nor to calculate the density estimate.  In our tests, we split our validation set into a ``val-train" set upon which we estimate the densities, and ``val-val" set which is only used for evaluation.
\end{itemize}

\paragraph{A note on variational inference}\label{sec:variational_inference}

While using variational methods to obtain probabilistic outputs from the neural network (such as ensembles or dropout) is possible, they are not required. They could be used to increase the numbers of unique predictions to fill out the density estimate, or to combine multiple trials in a Bayesian framework.  Given $T$ trials ($t$) returning $y_{t}$ for a single input $x$, we can combine the trials in a frequentist or Bayesian framework.  For each pixel $x$ and its set of trials $y_{t}\text{ for }t\in\left[1\dots T\right]$, we first must generate a hypothesis for the class of that pixel.  We do so by taking the argument of the maximum value of the sum of probabilities for that class, i.e.
\begin{equation}\label{eq:hypothesis}
    H\left(x\right) = \underset{c\in\left[1\dots d\right]}{\mathrm{argmax}}\left[\sum_{t=1}^{T} p\left(y_{t}|c_{t}\right)\right].
\end{equation}

Given that hypothesis, we can now combine the probabilities that the evidence from a given trial matches the class hypothesis.  We define the probability of a trial matching the hypothesis as
\begin{equation}\label{eq:ph}
    p_{H}\left(y_{t}\right) \equiv
    \begin{cases}
    p\left(y_{t}\right) & c_{t} = H\left(x\right)\\
    1 - p\left(y_{t}\right) & \text{otherwise}.
    \end{cases}
\end{equation}

Then, we can combine those probabilities. We note that each variational trial is \emph{dependent}, especially in the case of dropout, so using the product of these probabilities (as one would with independent trials) is inappropriate.  Instead, we can combine these trials using the geometric mean (following \cite{Nelson2017}) in a frequentist framework.  This leads to
\begin{equation}\label{eq:frequentist}
    p_{T, freq}\left(x|H\right) = \sqrt[T]{\prod_{t = 1}^{T} p_{H}\left(x\right)}.
\end{equation}

We can also use a Bayesian framework.  To do so, we can define some prior, $\tilde{p}$, either equally weighting all classes, or weighting by some known prevalence of the hypothesis class.  For the equal weighting case, $\tilde{p}=\frac{1}{d}$.  Then, we can combine these trials using the likelihood ratio.  The likelihood ratio is the ratio of the probability of the hypothesis being true given the evidence to the probability of the hypothesis being untrue given the evidence. Since $p_{H}$ is the probability of the prediction being correct, given input $x$, the likelihood ratio is then
\begin{equation}
    LR = \frac{p\left(H|x\right)}{p\left(\neg H|x\right)} = \frac{p_{H}\left(x\right)}{1-p_{H}\left(x\right)}.
\end{equation}

We can manipulate Bayes rule such that it can be written
\begin{equation}
    p\left(H|x\right) = \frac{p\left(x|H\right)p\left(H\right)}{p\left(x|H\right)p\left(H\right)+p\left(x|\neg H\right)p\left(\neg H\right)},
\end{equation}
where $\neg$ denotes the not operator. Dividing through by the term $p\left(y|H\right)p\left(H\right)$ and multiplying the top and bottom by $LR$, we find the update rule for a new trial is
\begin{equation}
    p_{t, Bayes}\left(x_{t}|H\right) = \frac{LR}{LR + \left(\frac{1}{p_{t-1}} - 1\right)},
\end{equation} where $t\in\left[1\dots T\right]$ and $p_{0}=\tilde{p}$ \cite{Brownlee1960}. This recurrence relation can be used to calculate $p_{T, Bayes}$.

With this method now formalized, we illustrate the method on two experiments.  The first, a toy problem, shows the method and its performance on a binary classifier with balanced classes. The second shows the method on semantic segmentation data, which has five different classes with extreme class imbalance, and on which we also perform variational trials.

\section{Experiments}

\subsection{Balanced, Binary Classification}

We begin our experiments with a toy problem, wherein we train a classifier to determine from which, of two, normal distributions a point has been drawn.

\paragraph{Data}

We generate a data set by first choosing a class of the set $C\in\left\{0, 1\right\}$ with equal probability. This class acts as the class label.  Then, the value of that point is drawn from a normal distribution such that $x\sim \mathcal{N}\left(\mu=3C, \sigma=1\right)$.

\paragraph{Method}\label{sec:method}

We train a multilayer perceptron to classify a point $x$ according to its class label $C$, providing the prediction $y$.  The multilayer perceptron that we build is comprised of two linear layers. The first expands the input into 16 nodes and is followed by rectified linear unit activation \cite{Goodfellow-et-al-2016}.  The second constricts from 16 nodes to 2 output nodes, and is followed by softmax activation.  This guarantees that the output is between zero and unity for each class.

We train the multilayer perceptron using the Adam \cite{Kingma2015} optimizer and minimizing the mean squared error (MSE) loss between the targets $C$ and predictions $y$\footnote{Categorical cross entropy is often used for classification, however we use MSE in this case to contend with label ambiguity due to the underlying distributions overlapping, which is akin to label noise.  MSE has been shown to be robust to label noise \cite{Ma2020}}.

We then perform calibration on a validation set as described above, using a histogram of 100 bins to estimate the densities $\rho$.  We evaluate our calibration by first creating a density estimation and calculating our ``Calibrated Score", $p$ from \cref{eq:prob_correct} on a new and completely unseen validation set (``val-val set").  We can also, for each histogram bin, retrieve the predicted labels $c$ and true labels $C$, and compute the true accuracy of those predictions.  If our calibration is successful, that true accuracy should be equal to the ``Calibrated Score" ($p$).

\paragraph{Results}

\begin{figure}
\centering
    \hspace*{-0.5cm}\scalebox{0.80}{\input{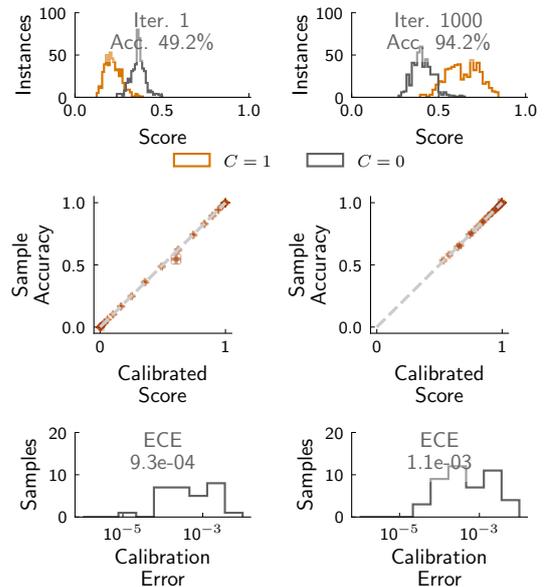}}
    \caption{Results on training and calibrating a binary classifier.  The scores (raw predicted outputs from a multilayer perceptron) as shown in row 1, the comparison of our calibrated score and the true accuracy of samples is shown in row 2, and the residuals between true accuracy and calibrated scores are shown in row 3.  The columns show different stages of training, varying from untrained to fully trained. Our classifier is within 1\% of perfectly calibrated in all cases.}
    \label{fig:binaryqqplot}
\end{figure}

Our calibration method works well for this binary classifier.  We show in \cref{fig:binaryqqplot} results for two checkpoints of our model, before and after training. Through each stage of training, our calibrated score is very close to the true accuracy for that histogram bin, in fact none of the differences between true accuracy and calibrated score are higher than $10^{-2}$. A common measure of calibration, the Expected Calibration Error (ECE) \cite{Nixon2019} is also very small, below $2\times 10^{-3}$ throughout training.

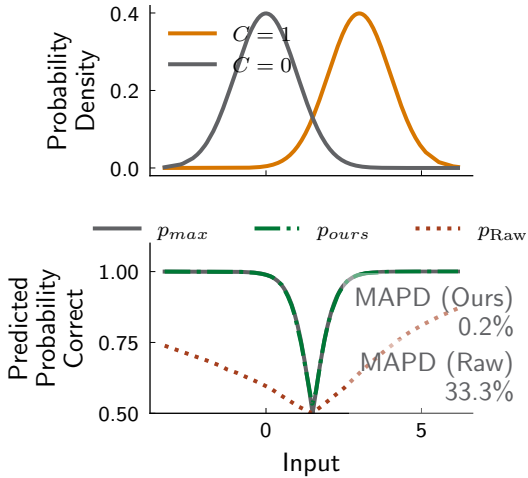
\begin{figure}
    \centering
    \input{best_probability.pgf}
    \caption{Probability density of two different classes given an input $x$, shown on the top panel.  The classes overlap near $x=1.5$. The maximum possible probability of determining from which class an input originates is shown in the bottom panel as $p_{max}$, and our calibrated score, $p_{ours}$, is very close to $p_{max}$. The raw output of the classifier is not calibrated well, shown as $p_{raw}$.}
    \label{fig:max_possible}
\end{figure}

In fact, because of the simplicity of this toy model, we can investigate the predictions and calibrated score against the true underlying probability distribution.  We can easily determine the ratio between the probability density of a single class and the marginal probability density over all classes given an input $x$.  A classifier cannot possibly predict with calibrated score above that ratio, because that is a measure of the ambiguity of the classification task.  In this toy problem, the maximum ratio of a single class divided by the overall marginal density should be close to one for all inputs $x$, except close to $1.5$, where the probability of $x$ coming from each of class 1 and 2 becomes close to 0.5.  We show this in \cref{fig:max_possible}, where the gray line shows the maximum possible calibrated score.  We see that the calibrated scores from our trained model and method match that line to within a mean absolute percentage difference (MAPD) of 0.1\%. This shows that the model is accurate in its predictions, and that our calibration is returning calibration scores very close to the actual ambiguity in the data set. We also show, in the red line, the raw neural network outputs given $x$, showing that these are not calibrated well.

\paragraph{Comparison to existing methods}
In the case of a general classifier, with no stipulations on architectural choices (such as the simple MLP we defined above), \emph{we are aware of no technique but ours to calculate the probability of predictions being correct}.  If we modify our choice of architecture by adding a dropout layer to the network, we can calculate the uncertainty in the model's predictions using a traditional variational inference technique \cite{Kendall2015}, which we call the baseline method.

In this framework, the variance of the raw neural network output over many trials is computed and used as the uncertainty.  While this is ambiguous regarding how that uncertainty relates to the probability that the prediction is correct, in the two class case, we can make some assumptions to generate this probability. Note that we \textbf{cannot} convert this variance to a probability in any but the two-class case.  Assuming that the raw neural network output is normally distributed, we can calculate the probability that a new sample would predict the same class as the current hypothesis.  First, we compute the sample mean $\mu$ and variance $\sigma$ given many trials of the classifier.  Then, the hypothesis is simply whether $\mu$ is greater than the threshold (0.5 for a softmax classifier).  The probability of that trial, $p_{t}$, is given by
\begin{equation}\label{eq:kendallpt}
    p_{t}\left(y_{t}\right) = 
    \begin{cases}
    1.0 - \phi\left(0.5 | \mu, \sigma\right) & \text{for }H=1 \\
    \phi\left(0.5 | \mu, \sigma\right) & \text{for }H=0
    \end{cases}
\end{equation}
where $\phi$ is the cumulative density function of a normal distribution. These trials can be combined by \cref{eq:frequentist}. This does not take into account the performance of the model, simply the model's belief in its own predictions.  Previous frameworks have no way to perform the former.

\begin{figure}
    \centering
    \input{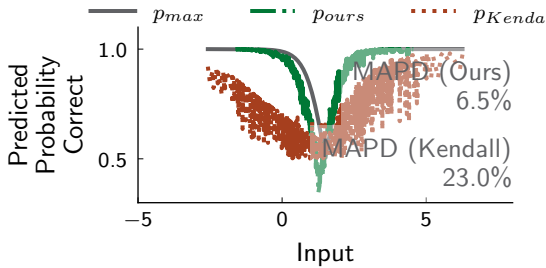}
    \caption{Calibrated probabilities using ours and the baseline method with a Monte Carlo dropout enabled MLP for a binary toy problem, compared with the maximum possible probability.  Our method is much closer to the maximum possible probability of being correct than baseline.  The baseline sometimes violates the maximum possible probability, because it does not evaluate the performance of the model at all, just its belief.}
    \label{fig:mcmc_calbration}
\end{figure}

To compare, we trained an MLP model of the same architecture as above, with the addition of a monte carlo dropout layer with dropout probability of 20\% after the input layer.  This model had reduced performance, compared to the above, even after twice the training effort (its accuracy was 87.6\%, whereas without dropout the accuracy is 94.2\%).  We then computed the probabilities of the prediction being correct using our framework and frequentist trial combination, using \cref{eq:ph} and \cref{eq:frequentist}.  We computed the probabilities using the baseline method and \cref{eq:kendallpt} and \cref{eq:frequentist}. We show the calibrated probabilities in \cref{fig:mcmc_calbration}, showing that our method is much closer to the maximum possible probability of being correct (a MAPD of 6.5\% compared to 23.0\%). We also show that, because the baseline method does not account for the performance of the model at all, it sometimes violated the maximum possible probability of being correct.

Overall, this investigation shows three benefits of our method compared to those in the literature, especially variational inference methods.  Our method does not require any architecture, and therefore may be used on the \emph{best performing model}, whereas others stipulate architecture designs which may affect performance. Our method calibrates to the model's belief and performance, and therefore cannot violate the ambiguity of the task itself, whereas other methods take no account of the performance of the model.  And finally, our method can be extended to multiple classes, whereas we are aware of no other variational methods that allow for the calculation of probabilities in the multinomial sense.

\subsection{Semantic segmentation of lithium aluminate micrographs}\label{sec:lialo2}

More difficult classification problems can also be attended with this technique. In this section, we show its use on an extremely imbalanced semantic segmentation problem.  This problem also is truly multinomial, having 5 different classes possible.

Previous work trained a SegNet model to perform semantic semgentation on micrographs of LiAlO$_{2}$ \cite{Pazdernik2020}.  This model returns an image which is the size of the input image in width and height, but consists of five channels, corresponding to each possible class in the image (Grain, Grain Boundary, Void, Impurity, and Precipitate).

In order to calculate \cref{eq:prob_correct} in a 5 class classification, we will need to estimate the density of correct and incorrect examples in 5 dimensions. In order to do so, we need to cover the 5 dimensional space, so we use variational inference of our semantic segmentation model to generate predicted values $y$ with more diversity.  We do this following the Bayesian SegNet technique laid out in \cite{Kendall2015}. We enable Dropout layers in $f$ during both training and inferences.  This makes our segmentation model probabilistic, ie. two different inferences of the model will generate different predictions. We denote these separate inference steps ``trials". We use the calibration method described in \cref{sec:method} to provide an estimate calibrated to the model's internal belief and its performance on validation data.

In practice, we found that, due to the high dimensional nature of our problem (5 classes), a histogram was not appropriate as a density estimator \cite{Scott2004}, and instead a k-Nearest Neighbors density estimator performed better \cite{Biau2011}.  We also found a weighted density estimate was required, as our majority class was $\sim200$ times more prevalent in the data set than the least prevalent of the minority classes.

The dropout layers in $f$ allow us to probe the model's belief of a prediction, and each of these trials constitutes new evidence for the probability of a prediction's correctness.
We can now combine this evidence following the frequentist or Bayesian methods from \cref{sec:variational_inference}. To avoid floating point arithmetic problems, we cast the product of all trial probabilities as the sum of logarithms.  We truncate possible values of the sum of logarithms to their relevant limits for floating 32-bit computation.  Finally, we can perform division by the number of trials and take the exponent to determine the probability estimate $\mathbf{\rho}\left(\vec{p}\right)$. Thus, the probability estimate is defined by
\begin{equation}
p_{T}\left(\vec{y}\right) =
\mathrm{exp}\left( \frac{1}{T}
\sum_{t=1}^{T}
\mathrm{log}\left(p\left(y_{t}\right)\right)\right)
\end{equation} This is exactly equal to the equation in \cref{eq:frequentist}, but safer from overflow and underflow errors when computing using floating point data types.

\subsection{Results}

To evaluate the performance of this estimate, we evaluate $f$ $25$ times on $40$ separate input images from our validation set.  We then separate $8$ images for use a val-val set for our calibration method ($\mathcal{V}$), leaving the other $32$ for our density estimate ($\mathcal{D}$).  We build KDTrees for every pixel in $D$, separated into different KDTrees for those which are correctly predicted and all predictions.  We use a k-Nearest Neighbors density estimator as in \cite{Biau2011}
\begin{equation}
    \rho_{\mathcal{Y}}\simeq\rho_{kNN}\left(x\right) = \frac{\sum_{i=1}^{k} i}{n V_{5} \sum_{i=1}^{k} d_{i}\left(x\right)^{5}}
\end{equation}
where $V_{5}$ is the volume of a 5-dimensional unit sphere, $n$ is the total number of points in the data set, and $d_{i}\left(x\right)$ is the distance from point $x$ to its $i$-th neighbor. We use $k=25$, as it performs well empirically while keeping computational effort reasonable. Higher numbers of $k$ would provide a better density estimate at the cost of computation time for the calibration.

\begin{figure}
    \centering
    \includegraphics{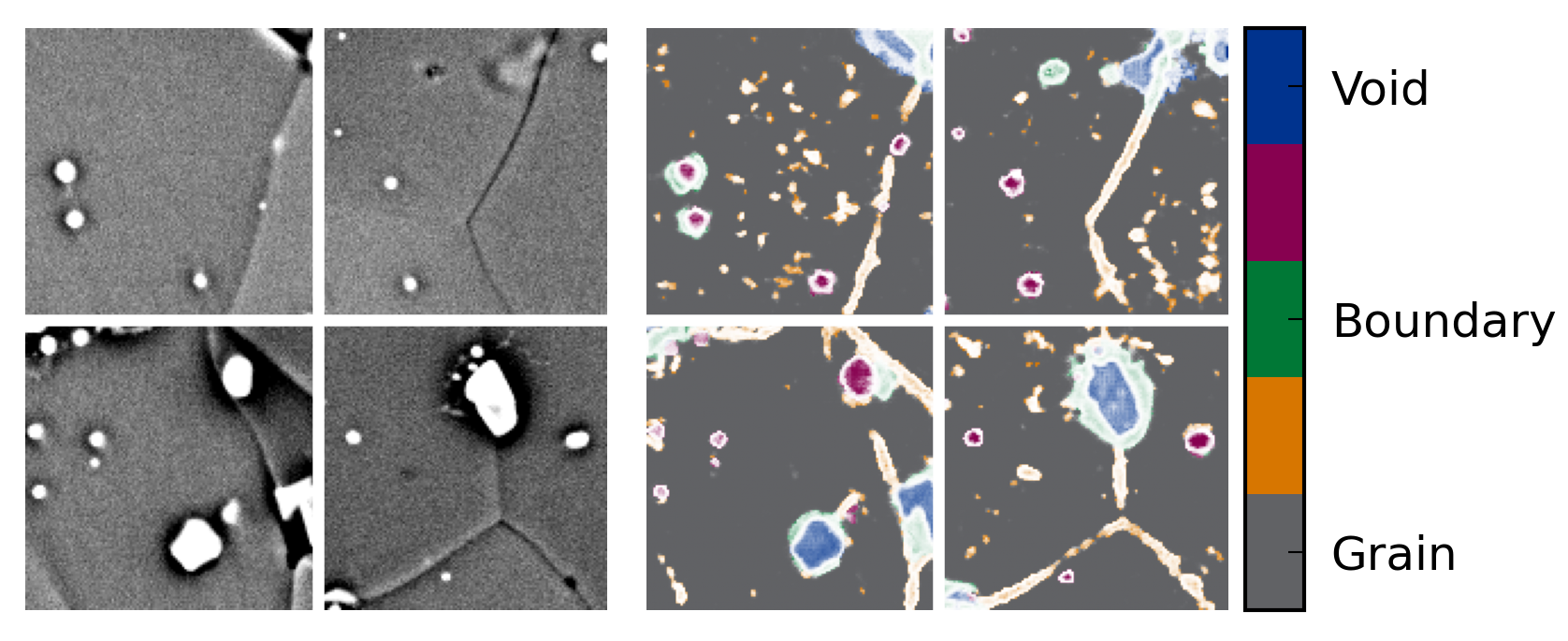}
    \caption{Segmentation of an input image with calibrated score displayed as transparency.  More transparent (whiter) areas are less certain in the classification of their pixels' class.}
    \label{fig:pred_seg}
\end{figure}

With the density estimates $\rho_{c}$ and $\rho_{a}$, we can calculate the probability $p$ for each pixel in $\mathcal{V}$.  The utility of this probability can be seen in \cref{fig:pred_seg}, where we show segmentation of different classes with the attendant uncertainty.  While we have no way to evaluate the true uncertainty in this segmentation task, the calibrated uncertainty shows that the model is uncertain around the boundaries between two classes, as expected.

\begin{figure}
    \centering
    \input{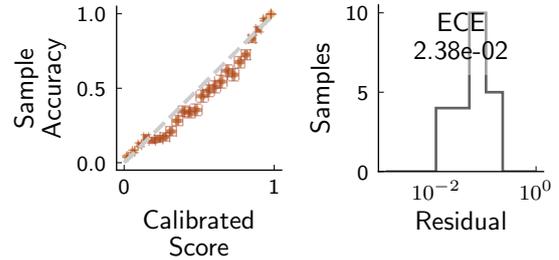}
    \caption{The sample accuracy of samples with similar calibrated scores show very close agreement with the desired calibration on segmented lithium aluminate images.  The residuals between the actual calibration and a perfect calibration are also small, with the expected calibration error less than 3\%.}
    \label{fig:qqplot}
\end{figure}

We can evaluate the calibration performance in the same way as the binary calibration example, by comparing the calibrated score to accuracy of samples with similar calibrated scores. We show results of this evaluation in \cref{fig:qqplot}, where the sample accuracy of samples with similar calibrated scores are plotted, and the residuals between the sample accuracy and calibrated scores. The expected calibration here is again very small, just over 2\%, showing that the calibration has been very successful.

\section{Conclusion}

We have developed a novel and flexible method for generating calibrated predictions from any classifier which returns a continuous value. This method uses predictions and labels from the data set to generate a comparison of the probability density of correct predictions to the probability density of total predictions.  This ratio, we show, is the probability that a new prediction will be correct given that the new data is from the same distribution as training and validation data.  We also describe the best practices for combining multiple trials into one prediction using variational inference and our framework.

This method outperforms other neural network uncertainty quantification methods because of its explicit calibration of both the classifier's \emph{belief} and \emph{performance}.  We have shown the success of this calibration method to less than 1\% expected calibration error on a balanced, binary classification problem.  Further, we showed less than 3\% expected calibration error on an extremely imbalanced semantic segmentation problem for labeling phases from lithium aluminate micrographs, the phase structure of which is an important question in nuclear science.

While this method is immediately applicable to many existing classifiers, we believe that extensions to 1) understand how to propagate this uncertainty through post-classifier analysis, and 2) explorations of methods for performing this calibration in a differentiable way during neural network training would be fruitful.

\bibliographystyle{ieeetr}
\bibliography{references.bib}
\end{document}

%% file: best_probability.pgf
\begingroup%
\makeatletter%
\begin{pgfpicture}%
\pgfpathrectangle{\pgfpointorigin}{\pgfqpoint{3.000000in}{2.781211in}}%
\pgfusepath{use as bounding box, clip}%
\begin{pgfscope}%
\pgfsetbuttcap%
\pgfsetmiterjoin%
\pgfsetlinewidth{0.000000pt}%
\definecolor{currentstroke}{rgb}{0.000000,0.000000,0.000000}%
\pgfsetstrokecolor{currentstroke}%
\pgfsetstrokeopacity{0.000000}%
\pgfsetdash{}{0pt}%
\pgfpathmoveto{\pgfqpoint{0.000000in}{0.000000in}}%
\pgfpathlineto{\pgfqpoint{3.000000in}{0.000000in}}%
\pgfpathlineto{\pgfqpoint{3.000000in}{2.781211in}}%
\pgfpathlineto{\pgfqpoint{0.000000in}{2.781211in}}%
\pgfpathclose%
\pgfusepath{}%
\end{pgfscope}%
\begin{pgfscope}%
\pgfsetbuttcap%
\pgfsetmiterjoin%
\pgfsetlinewidth{0.000000pt}%
\definecolor{currentstroke}{rgb}{0.000000,0.000000,0.000000}%
\pgfsetstrokecolor{currentstroke}%
\pgfsetstrokeopacity{0.000000}%
\pgfsetdash{}{0pt}%
\pgfpathmoveto{\pgfqpoint{0.886458in}{1.747200in}}%
\pgfpathlineto{\pgfqpoint{2.557815in}{1.747200in}}%
\pgfpathlineto{\pgfqpoint{2.557815in}{2.634794in}}%
\pgfpathlineto{\pgfqpoint{0.886458in}{2.634794in}}%
\pgfpathclose%
\pgfusepath{}%
\end{pgfscope}%
\begin{pgfscope}%
\pgfsetbuttcap%
\pgfsetroundjoin%
\definecolor{currentfill}{rgb}{0.000000,0.000000,0.000000}%
\pgfsetfillcolor{currentfill}%
\pgfsetlinewidth{0.250937pt}%
\definecolor{currentstroke}{rgb}{0.000000,0.000000,0.000000}%
\pgfsetstrokecolor{currentstroke}%
\pgfsetdash{}{0pt}%
\pgfsys@defobject{currentmarker}{\pgfqpoint{0.000000in}{0.000000in}}{\pgfqpoint{0.000000in}{0.027778in}}{%
\pgfpathmoveto{\pgfqpoint{0.000000in}{0.000000in}}%
\pgfpathlineto{\pgfqpoint{0.000000in}{0.027778in}}%
\pgfusepath{stroke,fill}%
}%
\begin{pgfscope}%
\pgfsys@transformshift{1.487284in}{1.747200in}%
\pgfsys@useobject{currentmarker}{}%
\end{pgfscope}%
\end{pgfscope}%
\begin{pgfscope}%
\pgfsetbuttcap%
\pgfsetroundjoin%
\definecolor{currentfill}{rgb}{0.000000,0.000000,0.000000}%
\pgfsetfillcolor{currentfill}%
\pgfsetlinewidth{0.250937pt}%
\definecolor{currentstroke}{rgb}{0.000000,0.000000,0.000000}%
\pgfsetstrokecolor{currentstroke}%
\pgfsetdash{}{0pt}%
\pgfsys@defobject{currentmarker}{\pgfqpoint{0.000000in}{0.000000in}}{\pgfqpoint{0.000000in}{0.027778in}}{%
\pgfpathmoveto{\pgfqpoint{0.000000in}{0.000000in}}%
\pgfpathlineto{\pgfqpoint{0.000000in}{0.027778in}}%
\pgfusepath{stroke,fill}%
}%
\begin{pgfscope}%
\pgfsys@transformshift{2.292633in}{1.747200in}%
\pgfsys@useobject{currentmarker}{}%
\end{pgfscope}%
\end{pgfscope}%
\begin{pgfscope}%
\pgfsetbuttcap%
\pgfsetroundjoin%
\definecolor{currentfill}{rgb}{0.000000,0.000000,0.000000}%
\pgfsetfillcolor{currentfill}%
\pgfsetlinewidth{0.250937pt}%
\definecolor{currentstroke}{rgb}{0.000000,0.000000,0.000000}%
\pgfsetstrokecolor{currentstroke}%
\pgfsetdash{}{0pt}%
\pgfsys@defobject{currentmarker}{\pgfqpoint{0.000000in}{0.000000in}}{\pgfqpoint{0.027778in}{0.000000in}}{%
\pgfpathmoveto{\pgfqpoint{0.000000in}{0.000000in}}%
\pgfpathlineto{\pgfqpoint{0.027778in}{0.000000in}}%
\pgfusepath{stroke,fill}%
}%
\begin{pgfscope}%
\pgfsys@transformshift{0.886458in}{1.787545in}%
\pgfsys@useobject{currentmarker}{}%
\end{pgfscope}%
\end{pgfscope}%
\begin{pgfscope}%
\definecolor{textcolor}{rgb}{0.000000,0.000000,0.000000}%
\pgfsetstrokecolor{textcolor}%
\pgfsetfillcolor{textcolor}%
\pgftext[x=0.680850in, y=1.747399in, left, base]{\color{textcolor}\sffamily\fontsize{8.330000}{9.996000}\selectfont 0.0}%
\end{pgfscope}%
\begin{pgfscope}%
\pgfsetbuttcap%
\pgfsetroundjoin%
\definecolor{currentfill}{rgb}{0.000000,0.000000,0.000000}%
\pgfsetfillcolor{currentfill}%
\pgfsetlinewidth{0.250937pt}%
\definecolor{currentstroke}{rgb}{0.000000,0.000000,0.000000}%
\pgfsetstrokecolor{currentstroke}%
\pgfsetdash{}{0pt}%
\pgfsys@defobject{currentmarker}{\pgfqpoint{0.000000in}{0.000000in}}{\pgfqpoint{0.027778in}{0.000000in}}{%
\pgfpathmoveto{\pgfqpoint{0.000000in}{0.000000in}}%
\pgfpathlineto{\pgfqpoint{0.027778in}{0.000000in}}%
\pgfusepath{stroke,fill}%
}%
\begin{pgfscope}%
\pgfsys@transformshift{0.886458in}{2.192067in}%
\pgfsys@useobject{currentmarker}{}%
\end{pgfscope}%
\end{pgfscope}%
\begin{pgfscope}%
\definecolor{textcolor}{rgb}{0.000000,0.000000,0.000000}%
\pgfsetstrokecolor{textcolor}%
\pgfsetfillcolor{textcolor}%
\pgftext[x=0.680850in, y=2.151921in, left, base]{\color{textcolor}\sffamily\fontsize{8.330000}{9.996000}\selectfont 0.2}%
\end{pgfscope}%
\begin{pgfscope}%
\pgfsetbuttcap%
\pgfsetroundjoin%
\definecolor{currentfill}{rgb}{0.000000,0.000000,0.000000}%
\pgfsetfillcolor{currentfill}%
\pgfsetlinewidth{0.250937pt}%
\definecolor{currentstroke}{rgb}{0.000000,0.000000,0.000000}%
\pgfsetstrokecolor{currentstroke}%
\pgfsetdash{}{0pt}%
\pgfsys@defobject{currentmarker}{\pgfqpoint{0.000000in}{0.000000in}}{\pgfqpoint{0.027778in}{0.000000in}}{%
\pgfpathmoveto{\pgfqpoint{0.000000in}{0.000000in}}%
\pgfpathlineto{\pgfqpoint{0.027778in}{0.000000in}}%
\pgfusepath{stroke,fill}%
}%
\begin{pgfscope}%
\pgfsys@transformshift{0.886458in}{2.596590in}%
\pgfsys@useobject{currentmarker}{}%
\end{pgfscope}%
\end{pgfscope}%
\begin{pgfscope}%
\definecolor{textcolor}{rgb}{0.000000,0.000000,0.000000}%
\pgfsetstrokecolor{textcolor}%
\pgfsetfillcolor{textcolor}%
\pgftext[x=0.680850in, y=2.556444in, left, base]{\color{textcolor}\sffamily\fontsize{8.330000}{9.996000}\selectfont 0.4}%
\end{pgfscope}%
\begin{pgfscope}%
\definecolor{textcolor}{rgb}{0.000000,0.000000,0.000000}%
\pgfsetstrokecolor{textcolor}%
\pgfsetfillcolor{textcolor}%
\pgftext[x=0.452683in, y=1.878011in, left, base,rotate=90.000000]{\color{textcolor}\sffamily\fontsize{10.000000}{12.000000}\selectfont Probability}%
\end{pgfscope}%
\begin{pgfscope}%
\definecolor{textcolor}{rgb}{0.000000,0.000000,0.000000}%
\pgfsetstrokecolor{textcolor}%
\pgfsetfillcolor{textcolor}%
\pgftext[x=0.596822in, y=1.975789in, left, base,rotate=90.000000]{\color{textcolor}\sffamily\fontsize{10.000000}{12.000000}\selectfont Density}%
\end{pgfscope}%
\begin{pgfscope}%
\pgfpathrectangle{\pgfqpoint{0.886458in}{1.747200in}}{\pgfqpoint{1.671357in}{0.887594in}}%
\pgfusepath{clip}%
\pgfsetrectcap%
\pgfsetroundjoin%
\pgfsetlinewidth{1.505625pt}%
\definecolor{currentstroke}{rgb}{0.843137,0.462745,0.000000}%
\pgfsetstrokecolor{currentstroke}%
\pgfsetdash{}{0pt}%
\pgfpathmoveto{\pgfqpoint{0.962429in}{1.787545in}}%
\pgfpathlineto{\pgfqpoint{1.400368in}{1.789081in}}%
\pgfpathlineto{\pgfqpoint{1.455555in}{1.792414in}}%
\pgfpathlineto{\pgfqpoint{1.493991in}{1.797693in}}%
\pgfpathlineto{\pgfqpoint{1.523357in}{1.804661in}}%
\pgfpathlineto{\pgfqpoint{1.549817in}{1.814189in}}%
\pgfpathlineto{\pgfqpoint{1.570770in}{1.824654in}}%
\pgfpathlineto{\pgfqpoint{1.592279in}{1.838775in}}%
\pgfpathlineto{\pgfqpoint{1.611789in}{1.855133in}}%
\pgfpathlineto{\pgfqpoint{1.632761in}{1.877105in}}%
\pgfpathlineto{\pgfqpoint{1.651133in}{1.900566in}}%
\pgfpathlineto{\pgfqpoint{1.671346in}{1.931358in}}%
\pgfpathlineto{\pgfqpoint{1.691517in}{1.967600in}}%
\pgfpathlineto{\pgfqpoint{1.711373in}{2.008768in}}%
\pgfpathlineto{\pgfqpoint{1.734889in}{2.064372in}}%
\pgfpathlineto{\pgfqpoint{1.764253in}{2.143014in}}%
\pgfpathlineto{\pgfqpoint{1.801956in}{2.254284in}}%
\pgfpathlineto{\pgfqpoint{1.858931in}{2.422360in}}%
\pgfpathlineto{\pgfqpoint{1.883374in}{2.484645in}}%
\pgfpathlineto{\pgfqpoint{1.902627in}{2.525910in}}%
\pgfpathlineto{\pgfqpoint{1.918348in}{2.553252in}}%
\pgfpathlineto{\pgfqpoint{1.933563in}{2.573517in}}%
\pgfpathlineto{\pgfqpoint{1.946191in}{2.585318in}}%
\pgfpathlineto{\pgfqpoint{1.957616in}{2.591875in}}%
\pgfpathlineto{\pgfqpoint{1.968246in}{2.594372in}}%
\pgfpathlineto{\pgfqpoint{1.978519in}{2.593449in}}%
\pgfpathlineto{\pgfqpoint{1.988319in}{2.589524in}}%
\pgfpathlineto{\pgfqpoint{1.998680in}{2.582189in}}%
\pgfpathlineto{\pgfqpoint{2.011440in}{2.568794in}}%
\pgfpathlineto{\pgfqpoint{2.026115in}{2.547745in}}%
\pgfpathlineto{\pgfqpoint{2.042927in}{2.516848in}}%
\pgfpathlineto{\pgfqpoint{2.061421in}{2.475597in}}%
\pgfpathlineto{\pgfqpoint{2.083951in}{2.417165in}}%
\pgfpathlineto{\pgfqpoint{2.117986in}{2.318113in}}%
\pgfpathlineto{\pgfqpoint{2.187903in}{2.112033in}}%
\pgfpathlineto{\pgfqpoint{2.216275in}{2.039430in}}%
\pgfpathlineto{\pgfqpoint{2.240149in}{1.986249in}}%
\pgfpathlineto{\pgfqpoint{2.262235in}{1.944013in}}%
\pgfpathlineto{\pgfqpoint{2.284942in}{1.907554in}}%
\pgfpathlineto{\pgfqpoint{2.314975in}{1.869500in}}%
\pgfpathlineto{\pgfqpoint{2.382725in}{1.818057in}}%
\pgfpathlineto{\pgfqpoint{2.456538in}{1.796047in}}%
\pgfpathlineto{\pgfqpoint{2.481844in}{1.792772in}}%
\pgfpathlineto{\pgfqpoint{2.481844in}{1.792772in}}%
\pgfusepath{stroke}%
\end{pgfscope}%
\begin{pgfscope}%
\pgfpathrectangle{\pgfqpoint{0.886458in}{1.747200in}}{\pgfqpoint{1.671357in}{0.887594in}}%
\pgfusepath{clip}%
\pgfsetrectcap%
\pgfsetroundjoin%
\pgfsetlinewidth{1.505625pt}%
\definecolor{currentstroke}{rgb}{0.380392,0.384314,0.396078}%
\pgfsetstrokecolor{currentstroke}%
\pgfsetdash{}{0pt}%
\pgfpathmoveto{\pgfqpoint{0.962429in}{1.791537in}}%
\pgfpathlineto{\pgfqpoint{1.043358in}{1.805631in}}%
\pgfpathlineto{\pgfqpoint{1.056243in}{1.810021in}}%
\pgfpathlineto{\pgfqpoint{1.108079in}{1.838040in}}%
\pgfpathlineto{\pgfqpoint{1.130769in}{1.857203in}}%
\pgfpathlineto{\pgfqpoint{1.148325in}{1.875684in}}%
\pgfpathlineto{\pgfqpoint{1.170964in}{1.904856in}}%
\pgfpathlineto{\pgfqpoint{1.181350in}{1.920416in}}%
\pgfpathlineto{\pgfqpoint{1.214403in}{1.979659in}}%
\pgfpathlineto{\pgfqpoint{1.238803in}{2.033037in}}%
\pgfpathlineto{\pgfqpoint{1.265149in}{2.099300in}}%
\pgfpathlineto{\pgfqpoint{1.298728in}{2.194216in}}%
\pgfpathlineto{\pgfqpoint{1.390959in}{2.462325in}}%
\pgfpathlineto{\pgfqpoint{1.411902in}{2.510749in}}%
\pgfpathlineto{\pgfqpoint{1.429885in}{2.544809in}}%
\pgfpathlineto{\pgfqpoint{1.445136in}{2.567293in}}%
\pgfpathlineto{\pgfqpoint{1.457980in}{2.581206in}}%
\pgfpathlineto{\pgfqpoint{1.469778in}{2.589699in}}%
\pgfpathlineto{\pgfqpoint{1.480466in}{2.593728in}}%
\pgfpathlineto{\pgfqpoint{1.489465in}{2.594376in}}%
\pgfpathlineto{\pgfqpoint{1.501221in}{2.591435in}}%
\pgfpathlineto{\pgfqpoint{1.511665in}{2.585259in}}%
\pgfpathlineto{\pgfqpoint{1.523357in}{2.574465in}}%
\pgfpathlineto{\pgfqpoint{1.536406in}{2.557784in}}%
\pgfpathlineto{\pgfqpoint{1.550296in}{2.535007in}}%
\pgfpathlineto{\pgfqpoint{1.568227in}{2.498732in}}%
\pgfpathlineto{\pgfqpoint{1.588073in}{2.450976in}}%
\pgfpathlineto{\pgfqpoint{1.614644in}{2.377821in}}%
\pgfpathlineto{\pgfqpoint{1.660811in}{2.239180in}}%
\pgfpathlineto{\pgfqpoint{1.700697in}{2.122981in}}%
\pgfpathlineto{\pgfqpoint{1.730439in}{2.045742in}}%
\pgfpathlineto{\pgfqpoint{1.754107in}{1.992153in}}%
\pgfpathlineto{\pgfqpoint{1.774096in}{1.952854in}}%
\pgfpathlineto{\pgfqpoint{1.796390in}{1.915511in}}%
\pgfpathlineto{\pgfqpoint{1.818053in}{1.885511in}}%
\pgfpathlineto{\pgfqpoint{1.836677in}{1.864290in}}%
\pgfpathlineto{\pgfqpoint{1.856934in}{1.845506in}}%
\pgfpathlineto{\pgfqpoint{1.879048in}{1.829444in}}%
\pgfpathlineto{\pgfqpoint{1.900909in}{1.817388in}}%
\pgfpathlineto{\pgfqpoint{1.925425in}{1.807500in}}%
\pgfpathlineto{\pgfqpoint{1.953548in}{1.799768in}}%
\pgfpathlineto{\pgfqpoint{1.986480in}{1.794168in}}%
\pgfpathlineto{\pgfqpoint{2.028471in}{1.790399in}}%
\pgfpathlineto{\pgfqpoint{2.088124in}{1.788313in}}%
\pgfpathlineto{\pgfqpoint{2.216562in}{1.787574in}}%
\pgfpathlineto{\pgfqpoint{2.481844in}{1.787545in}}%
\pgfpathlineto{\pgfqpoint{2.481844in}{1.787545in}}%
\pgfusepath{stroke}%
\end{pgfscope}%
\begin{pgfscope}%
\pgfsetrectcap%
\pgfsetmiterjoin%
\pgfsetlinewidth{0.501875pt}%
\definecolor{currentstroke}{rgb}{0.000000,0.000000,0.000000}%
\pgfsetstrokecolor{currentstroke}%
\pgfsetdash{}{0pt}%
\pgfpathmoveto{\pgfqpoint{0.886458in}{1.747200in}}%
\pgfpathlineto{\pgfqpoint{0.886458in}{2.634794in}}%
\pgfusepath{stroke}%
\end{pgfscope}%
\begin{pgfscope}%
\pgfsetrectcap%
\pgfsetmiterjoin%
\pgfsetlinewidth{0.501875pt}%
\definecolor{currentstroke}{rgb}{0.000000,0.000000,0.000000}%
\pgfsetstrokecolor{currentstroke}%
\pgfsetdash{}{0pt}%
\pgfpathmoveto{\pgfqpoint{0.886458in}{1.747200in}}%
\pgfpathlineto{\pgfqpoint{2.557815in}{1.747200in}}%
\pgfusepath{stroke}%
\end{pgfscope}%
\begin{pgfscope}%
\pgfsetrectcap%
\pgfsetroundjoin%
\pgfsetlinewidth{1.505625pt}%
\definecolor{currentstroke}{rgb}{0.843137,0.462745,0.000000}%
\pgfsetstrokecolor{currentstroke}%
\pgfsetdash{}{0pt}%
\pgfpathmoveto{\pgfqpoint{0.990583in}{2.490176in}}%
\pgfpathlineto{\pgfqpoint{1.221972in}{2.490176in}}%
\pgfusepath{stroke}%
\end{pgfscope}%
\begin{pgfscope}%
\definecolor{textcolor}{rgb}{0.000000,0.000000,0.000000}%
\pgfsetstrokecolor{textcolor}%
\pgfsetfillcolor{textcolor}%
\pgftext[x=1.314528in,y=2.449683in,left,base]{\color{textcolor}\sffamily\fontsize{8.330000}{9.996000}\selectfont \(\displaystyle C=1\)}%
\end{pgfscope}%
\begin{pgfscope}%
\pgfsetrectcap%
\pgfsetroundjoin%
\pgfsetlinewidth{1.505625pt}%
\definecolor{currentstroke}{rgb}{0.380392,0.384314,0.396078}%
\pgfsetstrokecolor{currentstroke}%
\pgfsetdash{}{0pt}%
\pgfpathmoveto{\pgfqpoint{0.990583in}{2.328898in}}%
\pgfpathlineto{\pgfqpoint{1.221972in}{2.328898in}}%
\pgfusepath{stroke}%
\end{pgfscope}%
\begin{pgfscope}%
\definecolor{textcolor}{rgb}{0.000000,0.000000,0.000000}%
\pgfsetstrokecolor{textcolor}%
\pgfsetfillcolor{textcolor}%
\pgftext[x=1.314528in,y=2.288405in,left,base]{\color{textcolor}\sffamily\fontsize{8.330000}{9.996000}\selectfont \(\displaystyle C=0\)}%
\end{pgfscope}%
\begin{pgfscope}%
\pgfsetbuttcap%
\pgfsetmiterjoin%
\pgfsetlinewidth{0.000000pt}%
\definecolor{currentstroke}{rgb}{0.000000,0.000000,0.000000}%
\pgfsetstrokecolor{currentstroke}%
\pgfsetstrokeopacity{0.000000}%
\pgfsetdash{}{0pt}%
\pgfpathmoveto{\pgfqpoint{0.886458in}{0.506944in}}%
\pgfpathlineto{\pgfqpoint{2.557815in}{0.506944in}}%
\pgfpathlineto{\pgfqpoint{2.557815in}{1.394539in}}%
\pgfpathlineto{\pgfqpoint{0.886458in}{1.394539in}}%
\pgfpathclose%
\pgfusepath{}%
\end{pgfscope}%
\begin{pgfscope}%
\pgfsetbuttcap%
\pgfsetroundjoin%
\definecolor{currentfill}{rgb}{0.000000,0.000000,0.000000}%
\pgfsetfillcolor{currentfill}%
\pgfsetlinewidth{0.250937pt}%
\definecolor{currentstroke}{rgb}{0.000000,0.000000,0.000000}%
\pgfsetstrokecolor{currentstroke}%
\pgfsetdash{}{0pt}%
\pgfsys@defobject{currentmarker}{\pgfqpoint{0.000000in}{0.000000in}}{\pgfqpoint{0.000000in}{0.027778in}}{%
\pgfpathmoveto{\pgfqpoint{0.000000in}{0.000000in}}%
\pgfpathlineto{\pgfqpoint{0.000000in}{0.027778in}}%
\pgfusepath{stroke,fill}%
}%
\begin{pgfscope}%
\pgfsys@transformshift{1.487284in}{0.506944in}%
\pgfsys@useobject{currentmarker}{}%
\end{pgfscope}%
\end{pgfscope}%
\begin{pgfscope}%
\definecolor{textcolor}{rgb}{0.000000,0.000000,0.000000}%
\pgfsetstrokecolor{textcolor}%
\pgfsetfillcolor{textcolor}%
\pgftext[x=1.487284in,y=0.458333in,,top]{\color{textcolor}\sffamily\fontsize{8.330000}{9.996000}\selectfont 0}%
\end{pgfscope}%
\begin{pgfscope}%
\pgfsetbuttcap%
\pgfsetroundjoin%
\definecolor{currentfill}{rgb}{0.000000,0.000000,0.000000}%
\pgfsetfillcolor{currentfill}%
\pgfsetlinewidth{0.250937pt}%
\definecolor{currentstroke}{rgb}{0.000000,0.000000,0.000000}%
\pgfsetstrokecolor{currentstroke}%
\pgfsetdash{}{0pt}%
\pgfsys@defobject{currentmarker}{\pgfqpoint{0.000000in}{0.000000in}}{\pgfqpoint{0.000000in}{0.027778in}}{%
\pgfpathmoveto{\pgfqpoint{0.000000in}{0.000000in}}%
\pgfpathlineto{\pgfqpoint{0.000000in}{0.027778in}}%
\pgfusepath{stroke,fill}%
}%
\begin{pgfscope}%
\pgfsys@transformshift{2.292633in}{0.506944in}%
\pgfsys@useobject{currentmarker}{}%
\end{pgfscope}%
\end{pgfscope}%
\begin{pgfscope}%
\definecolor{textcolor}{rgb}{0.000000,0.000000,0.000000}%
\pgfsetstrokecolor{textcolor}%
\pgfsetfillcolor{textcolor}%
\pgftext[x=2.292633in,y=0.458333in,,top]{\color{textcolor}\sffamily\fontsize{8.330000}{9.996000}\selectfont 5}%
\end{pgfscope}%
\begin{pgfscope}%
\definecolor{textcolor}{rgb}{0.000000,0.000000,0.000000}%
\pgfsetstrokecolor{textcolor}%
\pgfsetfillcolor{textcolor}%
\pgftext[x=1.722137in,y=0.300041in,,top]{\color{textcolor}\sffamily\fontsize{10.000000}{12.000000}\selectfont Input}%
\end{pgfscope}%
\begin{pgfscope}%
\pgfsetbuttcap%
\pgfsetroundjoin%
\definecolor{currentfill}{rgb}{0.000000,0.000000,0.000000}%
\pgfsetfillcolor{currentfill}%
\pgfsetlinewidth{0.250937pt}%
\definecolor{currentstroke}{rgb}{0.000000,0.000000,0.000000}%
\pgfsetstrokecolor{currentstroke}%
\pgfsetdash{}{0pt}%
\pgfsys@defobject{currentmarker}{\pgfqpoint{0.000000in}{0.000000in}}{\pgfqpoint{0.027778in}{0.000000in}}{%
\pgfpathmoveto{\pgfqpoint{0.000000in}{0.000000in}}%
\pgfpathlineto{\pgfqpoint{0.027778in}{0.000000in}}%
\pgfusepath{stroke,fill}%
}%
\begin{pgfscope}%
\pgfsys@transformshift{0.886458in}{0.506944in}%
\pgfsys@useobject{currentmarker}{}%
\end{pgfscope}%
\end{pgfscope}%
\begin{pgfscope}%
\definecolor{textcolor}{rgb}{0.000000,0.000000,0.000000}%
\pgfsetstrokecolor{textcolor}%
\pgfsetfillcolor{textcolor}%
\pgftext[x=0.619416in, y=0.466798in, left, base]{\color{textcolor}\sffamily\fontsize{8.330000}{9.996000}\selectfont 0.50}%
\end{pgfscope}%
\begin{pgfscope}%
\pgfsetbuttcap%
\pgfsetroundjoin%
\definecolor{currentfill}{rgb}{0.000000,0.000000,0.000000}%
\pgfsetfillcolor{currentfill}%
\pgfsetlinewidth{0.250937pt}%
\definecolor{currentstroke}{rgb}{0.000000,0.000000,0.000000}%
\pgfsetstrokecolor{currentstroke}%
\pgfsetdash{}{0pt}%
\pgfsys@defobject{currentmarker}{\pgfqpoint{0.000000in}{0.000000in}}{\pgfqpoint{0.027778in}{0.000000in}}{%
\pgfpathmoveto{\pgfqpoint{0.000000in}{0.000000in}}%
\pgfpathlineto{\pgfqpoint{0.027778in}{0.000000in}}%
\pgfusepath{stroke,fill}%
}%
\begin{pgfscope}%
\pgfsys@transformshift{0.886458in}{0.876775in}%
\pgfsys@useobject{currentmarker}{}%
\end{pgfscope}%
\end{pgfscope}%
\begin{pgfscope}%
\definecolor{textcolor}{rgb}{0.000000,0.000000,0.000000}%
\pgfsetstrokecolor{textcolor}%
\pgfsetfillcolor{textcolor}%
\pgftext[x=0.619416in, y=0.836629in, left, base]{\color{textcolor}\sffamily\fontsize{8.330000}{9.996000}\selectfont 0.75}%
\end{pgfscope}%
\begin{pgfscope}%
\pgfsetbuttcap%
\pgfsetroundjoin%
\definecolor{currentfill}{rgb}{0.000000,0.000000,0.000000}%
\pgfsetfillcolor{currentfill}%
\pgfsetlinewidth{0.250937pt}%
\definecolor{currentstroke}{rgb}{0.000000,0.000000,0.000000}%
\pgfsetstrokecolor{currentstroke}%
\pgfsetdash{}{0pt}%
\pgfsys@defobject{currentmarker}{\pgfqpoint{0.000000in}{0.000000in}}{\pgfqpoint{0.027778in}{0.000000in}}{%
\pgfpathmoveto{\pgfqpoint{0.000000in}{0.000000in}}%
\pgfpathlineto{\pgfqpoint{0.027778in}{0.000000in}}%
\pgfusepath{stroke,fill}%
}%
\begin{pgfscope}%
\pgfsys@transformshift{0.886458in}{1.246606in}%
\pgfsys@useobject{currentmarker}{}%
\end{pgfscope}%
\end{pgfscope}%
\begin{pgfscope}%
\definecolor{textcolor}{rgb}{0.000000,0.000000,0.000000}%
\pgfsetstrokecolor{textcolor}%
\pgfsetfillcolor{textcolor}%
\pgftext[x=0.619416in, y=1.206460in, left, base]{\color{textcolor}\sffamily\fontsize{8.330000}{9.996000}\selectfont 1.00}%
\end{pgfscope}%
\begin{pgfscope}%
\definecolor{textcolor}{rgb}{0.000000,0.000000,0.000000}%
\pgfsetstrokecolor{textcolor}%
\pgfsetfillcolor{textcolor}%
\pgftext[x=0.250166in, y=0.676644in, left, base,rotate=90.000000]{\color{textcolor}\sffamily\fontsize{10.000000}{12.000000}\selectfont Predicted}%
\end{pgfscope}%
\begin{pgfscope}%
\definecolor{textcolor}{rgb}{0.000000,0.000000,0.000000}%
\pgfsetstrokecolor{textcolor}%
\pgfsetfillcolor{textcolor}%
\pgftext[x=0.392777in, y=0.637756in, left, base,rotate=90.000000]{\color{textcolor}\sffamily\fontsize{10.000000}{12.000000}\selectfont Probability}%
\end{pgfscope}%
\begin{pgfscope}%
\definecolor{textcolor}{rgb}{0.000000,0.000000,0.000000}%
\pgfsetstrokecolor{textcolor}%
\pgfsetfillcolor{textcolor}%
\pgftext[x=0.536916in, y=0.739353in, left, base,rotate=90.000000]{\color{textcolor}\sffamily\fontsize{10.000000}{12.000000}\selectfont Correct}%
\end{pgfscope}%
\begin{pgfscope}%
\pgfpathrectangle{\pgfqpoint{0.886458in}{0.506944in}}{\pgfqpoint{1.671357in}{0.887594in}}%
\pgfusepath{clip}%
\pgfsetrectcap%
\pgfsetroundjoin%
\pgfsetlinewidth{1.505625pt}%
\definecolor{currentstroke}{rgb}{0.380392,0.384314,0.396078}%
\pgfsetstrokecolor{currentstroke}%
\pgfsetdash{}{0pt}%
\pgfpathmoveto{\pgfqpoint{0.962429in}{1.246605in}}%
\pgfpathlineto{\pgfqpoint{1.360885in}{1.245047in}}%
\pgfpathlineto{\pgfqpoint{1.422011in}{1.241750in}}%
\pgfpathlineto{\pgfqpoint{1.461304in}{1.236546in}}%
\pgfpathlineto{\pgfqpoint{1.489237in}{1.229758in}}%
\pgfpathlineto{\pgfqpoint{1.511665in}{1.221172in}}%
\pgfpathlineto{\pgfqpoint{1.528700in}{1.211898in}}%
\pgfpathlineto{\pgfqpoint{1.547050in}{1.198218in}}%
\pgfpathlineto{\pgfqpoint{1.561360in}{1.184064in}}%
\pgfpathlineto{\pgfqpoint{1.577073in}{1.163988in}}%
\pgfpathlineto{\pgfqpoint{1.590991in}{1.141279in}}%
\pgfpathlineto{\pgfqpoint{1.604083in}{1.114784in}}%
\pgfpathlineto{\pgfqpoint{1.619520in}{1.075939in}}%
\pgfpathlineto{\pgfqpoint{1.634491in}{1.029123in}}%
\pgfpathlineto{\pgfqpoint{1.651133in}{0.965128in}}%
\pgfpathlineto{\pgfqpoint{1.666484in}{0.894162in}}%
\pgfpathlineto{\pgfqpoint{1.685696in}{0.789397in}}%
\pgfpathlineto{\pgfqpoint{1.710265in}{0.633959in}}%
\pgfpathlineto{\pgfqpoint{1.728540in}{0.509346in}}%
\pgfpathlineto{\pgfqpoint{1.732942in}{0.534854in}}%
\pgfpathlineto{\pgfqpoint{1.765805in}{0.751666in}}%
\pgfpathlineto{\pgfqpoint{1.788644in}{0.880745in}}%
\pgfpathlineto{\pgfqpoint{1.806913in}{0.966270in}}%
\pgfpathlineto{\pgfqpoint{1.823609in}{1.030234in}}%
\pgfpathlineto{\pgfqpoint{1.839478in}{1.079343in}}%
\pgfpathlineto{\pgfqpoint{1.853147in}{1.113556in}}%
\pgfpathlineto{\pgfqpoint{1.870283in}{1.147476in}}%
\pgfpathlineto{\pgfqpoint{1.884424in}{1.169228in}}%
\pgfpathlineto{\pgfqpoint{1.900909in}{1.188887in}}%
\pgfpathlineto{\pgfqpoint{1.915008in}{1.201814in}}%
\pgfpathlineto{\pgfqpoint{1.933563in}{1.214620in}}%
\pgfpathlineto{\pgfqpoint{1.953548in}{1.224412in}}%
\pgfpathlineto{\pgfqpoint{1.975603in}{1.231814in}}%
\pgfpathlineto{\pgfqpoint{2.003505in}{1.237773in}}%
\pgfpathlineto{\pgfqpoint{2.039356in}{1.242063in}}%
\pgfpathlineto{\pgfqpoint{2.091200in}{1.244873in}}%
\pgfpathlineto{\pgfqpoint{2.182782in}{1.246291in}}%
\pgfpathlineto{\pgfqpoint{2.481844in}{1.246605in}}%
\pgfpathlineto{\pgfqpoint{2.481844in}{1.246605in}}%
\pgfusepath{stroke}%
\end{pgfscope}%
\begin{pgfscope}%
\pgfpathrectangle{\pgfqpoint{0.886458in}{0.506944in}}{\pgfqpoint{1.671357in}{0.887594in}}%
\pgfusepath{clip}%
\pgfsetbuttcap%
\pgfsetroundjoin%
\pgfsetlinewidth{1.505625pt}%
\definecolor{currentstroke}{rgb}{0.000000,0.470588,0.211765}%
\pgfsetstrokecolor{currentstroke}%
\pgfsetdash{{9.600000pt}{2.400000pt}{1.500000pt}{2.400000pt}}{0.000000pt}%
\pgfpathmoveto{\pgfqpoint{0.962429in}{1.246606in}}%
\pgfpathlineto{\pgfqpoint{1.358240in}{1.245076in}}%
\pgfpathlineto{\pgfqpoint{1.431039in}{1.240262in}}%
\pgfpathlineto{\pgfqpoint{1.469425in}{1.233943in}}%
\pgfpathlineto{\pgfqpoint{1.489465in}{1.229016in}}%
\pgfpathlineto{\pgfqpoint{1.511127in}{1.221911in}}%
\pgfpathlineto{\pgfqpoint{1.523357in}{1.215315in}}%
\pgfpathlineto{\pgfqpoint{1.531367in}{1.209485in}}%
\pgfpathlineto{\pgfqpoint{1.570770in}{1.176034in}}%
\pgfpathlineto{\pgfqpoint{1.600553in}{1.117528in}}%
\pgfpathlineto{\pgfqpoint{1.619520in}{1.071247in}}%
\pgfpathlineto{\pgfqpoint{1.639098in}{1.007107in}}%
\pgfpathlineto{\pgfqpoint{1.650386in}{0.954329in}}%
\pgfpathlineto{\pgfqpoint{1.670043in}{0.868467in}}%
\pgfpathlineto{\pgfqpoint{1.725210in}{0.556981in}}%
\pgfpathlineto{\pgfqpoint{1.727862in}{0.557409in}}%
\pgfpathlineto{\pgfqpoint{1.746006in}{0.620603in}}%
\pgfpathlineto{\pgfqpoint{1.756712in}{0.690792in}}%
\pgfpathlineto{\pgfqpoint{1.774096in}{0.800991in}}%
\pgfpathlineto{\pgfqpoint{1.792051in}{0.899021in}}%
\pgfpathlineto{\pgfqpoint{1.808725in}{0.974444in}}%
\pgfpathlineto{\pgfqpoint{1.827886in}{1.040965in}}%
\pgfpathlineto{\pgfqpoint{1.858931in}{1.122918in}}%
\pgfpathlineto{\pgfqpoint{1.865355in}{1.135392in}}%
\pgfpathlineto{\pgfqpoint{1.887868in}{1.172423in}}%
\pgfpathlineto{\pgfqpoint{1.906400in}{1.195692in}}%
\pgfpathlineto{\pgfqpoint{1.913554in}{1.201829in}}%
\pgfpathlineto{\pgfqpoint{1.945749in}{1.221146in}}%
\pgfpathlineto{\pgfqpoint{1.960213in}{1.227455in}}%
\pgfpathlineto{\pgfqpoint{2.004353in}{1.238578in}}%
\pgfpathlineto{\pgfqpoint{2.118863in}{1.245192in}}%
\pgfpathlineto{\pgfqpoint{2.262235in}{1.246606in}}%
\pgfpathlineto{\pgfqpoint{2.481844in}{1.246606in}}%
\pgfpathlineto{\pgfqpoint{2.481844in}{1.246606in}}%
\pgfusepath{stroke}%
\end{pgfscope}%
\begin{pgfscope}%
\pgfpathrectangle{\pgfqpoint{0.886458in}{0.506944in}}{\pgfqpoint{1.671357in}{0.887594in}}%
\pgfusepath{clip}%
\pgfsetbuttcap%
\pgfsetroundjoin%
\pgfsetlinewidth{1.505625pt}%
\definecolor{currentstroke}{rgb}{0.650980,0.247059,0.117647}%
\pgfsetstrokecolor{currentstroke}%
\pgfsetdash{{1.500000pt}{2.475000pt}}{0.000000pt}%
\pgfpathmoveto{\pgfqpoint{0.962429in}{0.859899in}}%
\pgfpathlineto{\pgfqpoint{1.181350in}{0.782350in}}%
\pgfpathlineto{\pgfqpoint{1.250301in}{0.755159in}}%
\pgfpathlineto{\pgfqpoint{1.286057in}{0.735743in}}%
\pgfpathlineto{\pgfqpoint{1.335836in}{0.718298in}}%
\pgfpathlineto{\pgfqpoint{1.452583in}{0.665842in}}%
\pgfpathlineto{\pgfqpoint{1.511665in}{0.636443in}}%
\pgfpathlineto{\pgfqpoint{1.550296in}{0.609534in}}%
\pgfpathlineto{\pgfqpoint{1.659163in}{0.540319in}}%
\pgfpathlineto{\pgfqpoint{1.725210in}{0.507429in}}%
\pgfpathlineto{\pgfqpoint{1.730439in}{0.510020in}}%
\pgfpathlineto{\pgfqpoint{1.765805in}{0.536993in}}%
\pgfpathlineto{\pgfqpoint{1.891284in}{0.638524in}}%
\pgfpathlineto{\pgfqpoint{1.977807in}{0.723202in}}%
\pgfpathlineto{\pgfqpoint{2.036881in}{0.777106in}}%
\pgfpathlineto{\pgfqpoint{2.099708in}{0.830978in}}%
\pgfpathlineto{\pgfqpoint{2.163818in}{0.881884in}}%
\pgfpathlineto{\pgfqpoint{2.223692in}{0.925483in}}%
\pgfpathlineto{\pgfqpoint{2.284942in}{0.961074in}}%
\pgfpathlineto{\pgfqpoint{2.314975in}{0.977393in}}%
\pgfpathlineto{\pgfqpoint{2.382725in}{1.011526in}}%
\pgfpathlineto{\pgfqpoint{2.481844in}{1.055036in}}%
\pgfpathlineto{\pgfqpoint{2.481844in}{1.055036in}}%
\pgfusepath{stroke}%
\end{pgfscope}%
\begin{pgfscope}%
\pgfsetrectcap%
\pgfsetmiterjoin%
\pgfsetlinewidth{0.501875pt}%
\definecolor{currentstroke}{rgb}{0.000000,0.000000,0.000000}%
\pgfsetstrokecolor{currentstroke}%
\pgfsetdash{}{0pt}%
\pgfpathmoveto{\pgfqpoint{0.886458in}{0.506944in}}%
\pgfpathlineto{\pgfqpoint{0.886458in}{1.394539in}}%
\pgfusepath{stroke}%
\end{pgfscope}%
\begin{pgfscope}%
\pgfsetrectcap%
\pgfsetmiterjoin%
\pgfsetlinewidth{0.501875pt}%
\definecolor{currentstroke}{rgb}{0.000000,0.000000,0.000000}%
\pgfsetstrokecolor{currentstroke}%
\pgfsetdash{}{0pt}%
\pgfpathmoveto{\pgfqpoint{0.886458in}{0.506944in}}%
\pgfpathlineto{\pgfqpoint{2.557815in}{0.506944in}}%
\pgfusepath{stroke}%
\end{pgfscope}%
\begin{pgfscope}%
\pgfsetbuttcap%
\pgfsetmiterjoin%
\definecolor{currentfill}{rgb}{1.000000,1.000000,1.000000}%
\pgfsetfillcolor{currentfill}%
\pgfsetfillopacity{0.400000}%
\pgfsetlinewidth{0.000000pt}%
\definecolor{currentstroke}{rgb}{0.000000,0.000000,0.000000}%
\pgfsetstrokecolor{currentstroke}%
\pgfsetstrokeopacity{0.400000}%
\pgfsetdash{}{0pt}%
\pgfpathmoveto{\pgfqpoint{1.888066in}{0.835585in}}%
\pgfpathlineto{\pgfqpoint{2.831399in}{0.835585in}}%
\pgfpathlineto{\pgfqpoint{2.831399in}{1.235973in}}%
\pgfpathlineto{\pgfqpoint{1.888066in}{1.235973in}}%
\pgfpathclose%
\pgfusepath{fill}%
\end{pgfscope}%
\begin{pgfscope}%
\definecolor{textcolor}{rgb}{0.380392,0.384314,0.396078}%
\pgfsetstrokecolor{textcolor}%
\pgfsetfillcolor{textcolor}%
\pgftext[x=1.943621in, y=1.068474in, left, base]{\color{textcolor}\sffamily\fontsize{10.000000}{12.000000}\selectfont MAPD (Ours)}%
\end{pgfscope}%
\begin{pgfscope}%
\definecolor{textcolor}{rgb}{0.380392,0.384314,0.396078}%
\pgfsetstrokecolor{textcolor}%
\pgfsetfillcolor{textcolor}%
\pgftext[x=2.482649in, y=0.918085in, left, base]{\color{textcolor}\sffamily\fontsize{10.000000}{12.000000}\selectfont 0.2\%}%
\end{pgfscope}%
\begin{pgfscope}%
\pgfsetbuttcap%
\pgfsetmiterjoin%
\definecolor{currentfill}{rgb}{1.000000,1.000000,1.000000}%
\pgfsetfillcolor{currentfill}%
\pgfsetfillopacity{0.400000}%
\pgfsetlinewidth{0.000000pt}%
\definecolor{currentstroke}{rgb}{0.000000,0.000000,0.000000}%
\pgfsetstrokecolor{currentstroke}%
\pgfsetstrokeopacity{0.400000}%
\pgfsetdash{}{0pt}%
\pgfpathmoveto{\pgfqpoint{1.915288in}{0.495341in}}%
\pgfpathlineto{\pgfqpoint{2.831399in}{0.495341in}}%
\pgfpathlineto{\pgfqpoint{2.831399in}{0.895729in}}%
\pgfpathlineto{\pgfqpoint{1.915288in}{0.895729in}}%
\pgfpathclose%
\pgfusepath{fill}%
\end{pgfscope}%
\begin{pgfscope}%
\definecolor{textcolor}{rgb}{0.380392,0.384314,0.396078}%
\pgfsetstrokecolor{textcolor}%
\pgfsetfillcolor{textcolor}%
\pgftext[x=1.970843in, y=0.728229in, left, base]{\color{textcolor}\sffamily\fontsize{10.000000}{12.000000}\selectfont MAPD (Raw)}%
\end{pgfscope}%
\begin{pgfscope}%
\definecolor{textcolor}{rgb}{0.380392,0.384314,0.396078}%
\pgfsetstrokecolor{textcolor}%
\pgfsetfillcolor{textcolor}%
\pgftext[x=2.413204in, y=0.577840in, left, base]{\color{textcolor}\sffamily\fontsize{10.000000}{12.000000}\selectfont 33.3\%}%
\end{pgfscope}%
\begin{pgfscope}%
\pgfsetrectcap%
\pgfsetroundjoin%
\pgfsetlinewidth{1.505625pt}%
\definecolor{currentstroke}{rgb}{0.380392,0.384314,0.396078}%
\pgfsetstrokecolor{currentstroke}%
\pgfsetdash{}{0pt}%
\pgfpathmoveto{\pgfqpoint{0.600671in}{1.471819in}}%
\pgfpathlineto{\pgfqpoint{0.832060in}{1.471819in}}%
\pgfusepath{stroke}%
\end{pgfscope}%
\begin{pgfscope}%
\definecolor{textcolor}{rgb}{0.000000,0.000000,0.000000}%
\pgfsetstrokecolor{textcolor}%
\pgfsetfillcolor{textcolor}%
\pgftext[x=0.924615in,y=1.431326in,left,base]{\color{textcolor}\sffamily\fontsize{8.330000}{9.996000}\selectfont \(\displaystyle p_{max}\)}%
\end{pgfscope}%
\begin{pgfscope}%
\pgfsetbuttcap%
\pgfsetroundjoin%
\pgfsetlinewidth{1.505625pt}%
\definecolor{currentstroke}{rgb}{0.000000,0.470588,0.211765}%
\pgfsetstrokecolor{currentstroke}%
\pgfsetdash{{9.600000pt}{2.400000pt}{1.500000pt}{2.400000pt}}{0.000000pt}%
\pgfpathmoveto{\pgfqpoint{1.427921in}{1.471819in}}%
\pgfpathlineto{\pgfqpoint{1.659310in}{1.471819in}}%
\pgfusepath{stroke}%
\end{pgfscope}%
\begin{pgfscope}%
\definecolor{textcolor}{rgb}{0.000000,0.000000,0.000000}%
\pgfsetstrokecolor{textcolor}%
\pgfsetfillcolor{textcolor}%
\pgftext[x=1.751866in,y=1.431326in,left,base]{\color{textcolor}\sffamily\fontsize{8.330000}{9.996000}\selectfont \(\displaystyle p_{ours}\)}%
\end{pgfscope}%
\begin{pgfscope}%
\pgfsetbuttcap%
\pgfsetroundjoin%
\pgfsetlinewidth{1.505625pt}%
\definecolor{currentstroke}{rgb}{0.650980,0.247059,0.117647}%
\pgfsetstrokecolor{currentstroke}%
\pgfsetdash{{1.500000pt}{2.475000pt}}{0.000000pt}%
\pgfpathmoveto{\pgfqpoint{2.261836in}{1.471819in}}%
\pgfpathlineto{\pgfqpoint{2.493225in}{1.471819in}}%
\pgfusepath{stroke}%
\end{pgfscope}%
\begin{pgfscope}%
\definecolor{textcolor}{rgb}{0.000000,0.000000,0.000000}%
\pgfsetstrokecolor{textcolor}%
\pgfsetfillcolor{textcolor}%
\pgftext[x=2.585781in,y=1.431326in,left,base]{\color{textcolor}\sffamily\fontsize{8.330000}{9.996000}\selectfont \(\displaystyle p_{\mathrm{Raw}}\)}%
\end{pgfscope}%
\end{pgfpicture}%
\makeatother%
\endgroup%